\documentclass[conference]{IEEEtran}
\IEEEoverridecommandlockouts
\usepackage{cite}
\usepackage{amsmath,amssymb,amsfonts}
\usepackage{bm}
\usepackage{algorithmic}
\usepackage{graphicx}
\usepackage{textcomp}
\usepackage{xcolor}
\def\BibTeX{{\rm B\kern-.05em{\sc i\kern-.025em b}\kern-.08em
    T\kern-.1667em\lower.7ex\hbox{E}\kern-.125emX}}
    
\newcommand{\E}{\operatorname{E}}
\newcommand{\Var}{\operatorname{Var}}
\newcommand{\Cov}{\operatorname{Cov}} 
\newcommand{\Gmax}{{G_\mathrm{max}}}

\begin{document}

\title{MemSE: Fast MSE Prediction for Noisy Memristor-Based DNN Accelerators
\thanks{This work was supported by an IVADO grant (PRF-2019-4784991664) and by the Samuel-de-Champlain program.}
}

% List of authors: Jonathan Kern, Sébastien Henwood, Gonçalo Mordido, Elsa Dupraz, Abdeldjalil Aïssa-El-Bey, Yvon Savaria, and François Leduc-Primeau
\author{
  \IEEEauthorblockN{Jonathan Kern\textsuperscript{1,2}, Sébastien Henwood\textsuperscript{1}, Gonçalo Mordido\textsuperscript{1,3}, Elsa Dupraz\textsuperscript{2},\\ Abdeldjalil Aïssa-El-Bey\textsuperscript{2}, Yvon Savaria\textsuperscript{1}, and François Leduc-Primeau\textsuperscript{1}}
 \IEEEauthorblockA{\textsuperscript{1}Department of Electrical Engineering, Polytechnique Montreal, Montreal, QC, Canada\\
 \textsuperscript{2}IMT Atlantique, CNRS UMR 6285, Lab-STICC, Brest, France\\
 \textsuperscript{3}Mila - Quebec AI Institute, Montreal, QC, Canada}
}

\maketitle

\begin{abstract}
Memristors enable the computation of matrix-vector multiplications (MVM) in memory and, therefore, show great potential in highly increasing the energy efficiency of deep neural network (DNN) inference accelerators. However, computations in memristors suffer from hardware non-idealities and are subject to different sources of noise that may negatively impact system performance. In this work, we theoretically analyze the mean squared error of DNNs that use memristor crossbars to compute MVM. We take into account both the quantization noise, due to the necessity of reducing the DNN model size, and the programming noise, stemming from the variability during the programming of the memristance value. 
Simulations on pre-trained DNN models showcase the accuracy of the analytical prediction. 
Furthermore
the proposed method is almost two order of magnitude faster than Monte-Carlo simulation, thus making it possible to optimize the implementation parameters to achieve minimal error for a given power constraint.
%The power consumption of the memristors is taken into account to formulate an optimization problem for minimizing the mean squared error of the neural network output for a given power constraint. 
\end{abstract}

%\begin{IEEEkeywords} %TODO: consider removing to save space unless the conf requires it
%in-memory computing, memristors, neural networks, energy efficiency
%\end{IEEEkeywords}

\section{Introduction}
Energy consumption represents one of the most important design objectives for deep neural network (DNN) accelerators, in particular, because it enables low-latency on-device processing. % Removing some references because the 4-page limit includes references.
%Over the last decades, the energy consumption of electronic systems has been continuously increasing~\cite{strubell_energy_2019}. % FLP: I do not think this is true in general. You probably mean for DNN models. (In any case I suggest to shorten this part.)
%As new applications emerge, the need for low-energy embedded inference engines is becoming of utmost importance, particularly in neural networks~\cite{sze2017efficient}.  
The main source of energy consumption in most accelerators is due to data movement~\cite{pedram_dark_2017}, specifically retrieving data from memory and delivering it to processing units. To bypass this problem, in-memory computing techniques show great promise in terms of energy efficiency by directly processing the data in memory. Particularly, memristors are an emerging technology well-suited for neural networks, which allow performing computations, such as dot products, in memory~\cite{sebastian_memory_2020}. 
Despite the energy benefits, programming the values of the conductance in memristor crossbars is an inexact process subject to noise~\cite{chen_variability_2011}. 
For instance, existing hardware implementations report precisions from 2 bits~\cite{perez_toward_2019} to 7.5 bits~\cite{hu_memristor-based_2018} per memristor. While additional techniques such as bit-slicing~\cite{diware_unbalanced_2021} may be leveraged to increase precision, this comes at the cost of increased area and energy usage.

DNNs have been shown to be robust to noise affecting the weights, although the amount of noise must be designed carefully to satisfy accuracy constraints~\cite{henwood_layerwise_2020,hacene_training_2019,hirtzlin_outstanding_2019}. 
Over the past few years, implementing neural networks using memristors has attracted a lot of attention~\cite{li_efficient_2018,hu_memristor-based_2018}. However, recent works focus mostly on the hardware architecture design and experimental results, neglecting theoretical analyses. 
One exception is \cite{dupraz_power-efficient_2021}, which presented a theoretical framework for DNN inference on memristors based on tracking second-order statistical moments. However, they used a crossbar model based on passive summing circuits, rather than active ones as in this paper, and did not consider quantization in the conductance values. Furthermore, the accuracy of the method was only verified on very small DNNs.

In this work, we analytically study neural network inference on memristor crossbars. To this end, we provide theoretical computations, which take into account practical implementation non-idealities, such as programming noise and quantization errors. Using these computations, we predict the mean squared error (MSE) at the output of the final layer of a neural network, depending on the network's parameters and scaling factors. Theoretical formulas are also provided to compute the power usage of the memristors crossbars depending on the scaling factors. Finally, we combine these analyses to formulate an optimization problem to minimize the MSE for a desired power usage. Lastly, simulations are performed to verify the accuracy of the theoretical analysis.% as well as examples of the optimization problem.

\section{Models}

\subsection{Memristor crossbar model}

%In this work, we consider a 1T1R crossbar implementation since it facilitates the programming of the devices~\cite{li_analogue_2018, li_memristive_2021}. 
%Several algorithms are possible for programming the conductance values, each presenting different advantages and drawbacks in term of accuracy and energy consumption\cite{puglisi_novel_2015,perez_toward_2019,perez-avila_behavioral_2020}.

%Given a weight matrix $W$ and an input vector $x$, we want to compute the following matrix-vector multiplication:
%\begin{equation}
%    z_j = \sum_{i=1}^L  w_{i,j} x_i \, ,
%\end{equation}

%
%In the ideal case, and following the memristor models presented by Hu \textit{et al.}~\cite{hu_dot-product_2016} and Li \textit{et al.}~\cite{li_efficient_2018}, the previous computation is:

\begin{figure}[t]
\centerline{\includegraphics[width=0.6\linewidth]{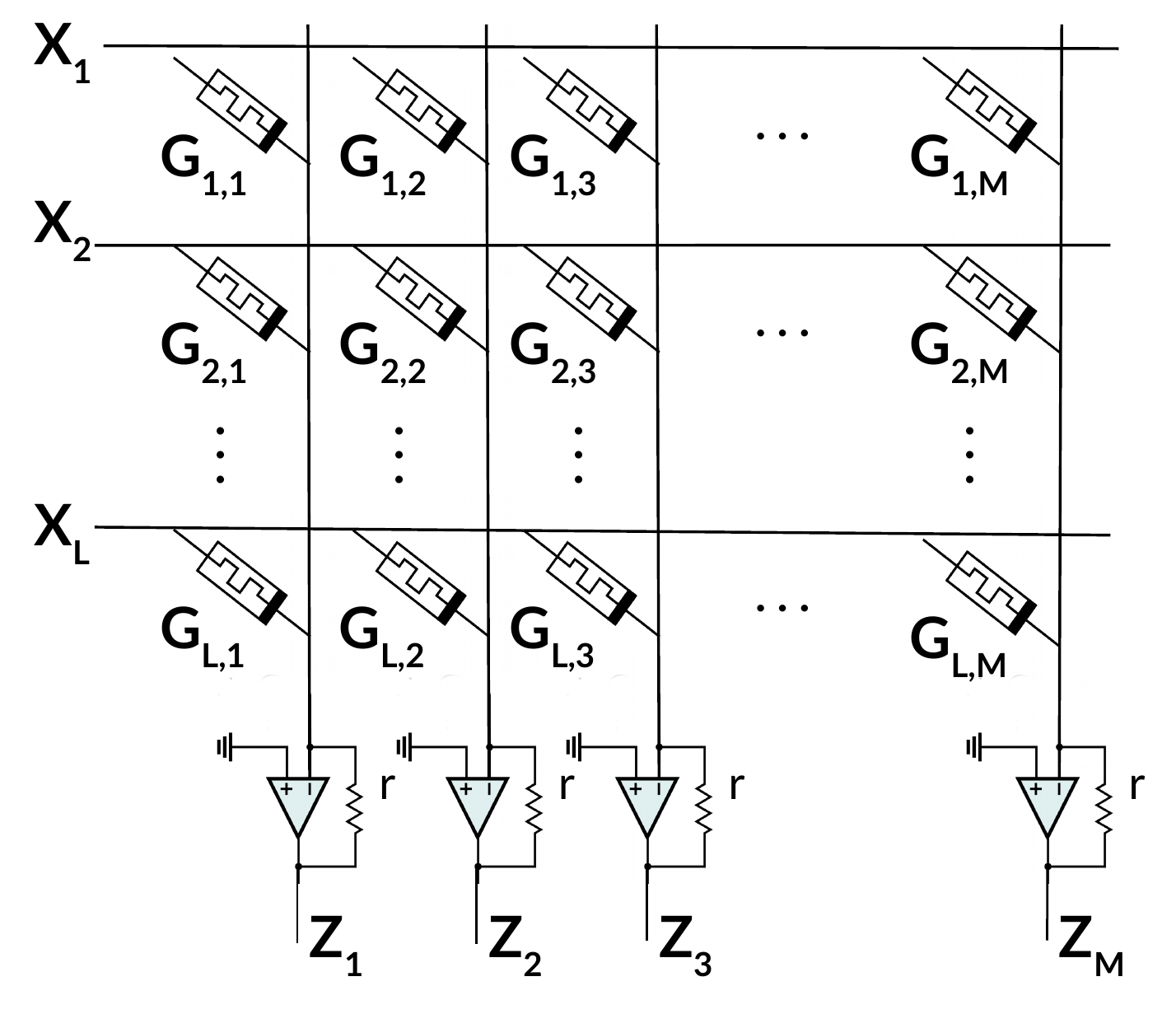}}
\caption{Memristor crossbar architecture for matrix-vector multiplication.}
\label{fig:schema}
\end{figure}

Figure~\ref{fig:schema} illustrates the architecture of the considered memristor crossbar. In accordance with Ohm's Law and Kirchoff's Law, the conductance at each node is multiplied with the input voltage of the row and these products are then summed along the column. Finally, a transimpedance amplifier (TIA) converts the current into a voltage at end of each column.
In the ideal case, the output of the $j$-th column is thus given by $z_j = r \sum_{i=1}^L g_{i,j} x_i$, where $x_i$ is the voltage at the input of row $i$, $g_{i,j}$ is the conductance of the memristor at row $i$ and column $j$, and r is the feedback resistance of the TIA.

However, several practical issues may cause the actual computation %done in a memristor crossbar 
to differ from the aforementioned ideal case. 
Specifically, values may be affected by fabrication variations and noise during programming~\cite{liu_memristor-based_2018,perez-avila_behavioral_2020,milo_multilevel_2019} as well as quantization errors.
With these practical constraints in mind, we consider that the memristors have a conductance ranging from $0$ to $G_\text{max}$, and divide this range into $N$ possible values. $G_\text{max}$ is chosen depending on the desired trade-off between accuracy and power consumption and only needs to be inferior to the maximum  physical conductance value. 
We denote the resulting resolution as $\Delta = \frac{G_{\max}}{N}$.
The programmed conductance values can then be represented as random variables $G_{i,j}$, which may be decomposed as
\begin{equation}
    G_{i,j} = g_{i,j}+\delta^q_{i,j}+\epsilon^v_{i,j} \, ,
\end{equation}
where $g_{i,j}$ is the desired value, $\delta^q_{i,j}$ is the quantization error, and $\epsilon^v_{i,j}$ is the noise due to variability in conductance programming.
We denote by $\sigma_v^2$ the variance of $\epsilon^v$. 
In practice, there can be different $\sigma_v$ for each possible memristor value, but here, to simplify the notations and computations, we consider that $\sigma_v$ is constant for all $N$ possible conductance values. The analysis proposed in this paper remains valid in the case where $\sigma_v$ is allowed to depend on the conductance value.

Since memristors can only store positive values, each weight $w_{i,j}$ 
is decomposed as $w_{i,j}=w_{i,j}^+-w_{i,j}^-$, where $w_{i,j}^+$ and $w_{i,j}^-$ store the positive and negative value of $w_{i,j}$, respectively. Then $w^+$ and $w^-$ are converted to the conductance $g_{i,j}^+$ and $g_{i,j}^-$. 
The matrix-vector multiplication can then be realized as
\begin{equation}
\label{eq:mvm_pos_neg}
    Z_j = \sum_{i=1}^L r G^+_{i,j} X_i-\sum_{i=1}^L r G^-_{i,j} X_i \, ,
\end{equation}
where $Z_j$, $G^+_{i,j}$, $G^-_{i,j}$, and $X_i$ are random variables.

%follows a uniform law of variance $\mathbb{V}[\delta^q]=\frac{\Delta^2}{12}$ where $\Delta=\frac{G_\text{max}}{N}$ is the quantization resolution. 

%Different models are possible for the variability error $\epsilon^v$ but it is possible to consider it Gaussian\cite{}. 

\subsection{Computation model}
\label{sec:comp_model}

For our theoretical analysis, we consider a neural network composed of convolutional, average pooling, and linear layers as well as differentiable activation functions. For simplicity, we consider that all convolutional layers are converted to linear layers. Moreover, batch normalization is not considered but could be easily incorporated into our analysis. %We note that the analysis is performed iteratively over the neural network layers. 

Because of the range of conductance possible, the matrix weights $w_{i,j}$ is scaled by a factor $c=\frac{G_\text{max}}{W_\text{max}}$, such that
\begin{equation}
\label{eq:conductance_values}
g_{i,j} = c w_{i,j}.
\end{equation}
We then divide the result of the memristor computations by this same factor $c$.
Denoting by $\Bar{g_{i,j}} = c w_{i,j} + \delta^q(c w_{i,j})$ the quantized version of $g$, where $\delta^q(c w_{i,j})$ is the (deterministic) quantization error, it should be noted that $\delta^q(c w_{i,j})=c\delta^q(w_{i,j})$. Therefore $\frac{\Bar{g_{i,j}}}{c} =  w_{i,j} + \delta^q(w_{i,j})$.

%
%and we note $\delta^q(w)$ the quantized error in the case of $c=1$(no scaling) then it should be noted that $\Bar{g} = c (w + \delta^q(w))$.

% More specifically, considering a given $w$, let us first define $g_1 = c_1 w$ and $g_2 = c_2w$, where $c_1 = \frac{\Gmax_1}{W_\text{max}}$ and $c_2 = \frac{\Gmax_2}{W_\text{max}}$. Then, quantizing $g_1$ and $g_2$ leads to $G_1 = g_1+\delta_1$ and $G_2 = g_2+\delta_2$, respectively, with the former having a quantization resolution of $\Delta_1 = \frac{\Gmax_1}{N}$ and the latter $\Delta_2 = \frac{\Gmax_2}{N}$. In the end, we have:
% \begin{equation}
%     \frac{\delta_1}{c_1} = \frac{\delta_2}{c_2}.
% \end{equation}
% %
% \todo{This sentence's text and equations need revision}
%Therefore for given values $w$ and $c$ such that $G = cw+\delta_w(c)$ we write $\delta_w = \frac{\delta_w(c)}{c}$.

For a given linear layer, the conductance values $g_{i,j}$ are computed following \eqref{eq:conductance_values} and uniformly quantized over the conductance range $[0,\Gmax]$. Then, the memristors products $\tilde{Z_j}^+ = \sum_{i=1}^L r G_{i,j}^+ x_i$ and $\tilde{Z_j}^- = \sum_{i=1}^L r G_{i,j}^- x_i$ are computed. The difference $\tilde{Z_j} = \tilde{Z_j}^+ - \tilde{Z_j}^-$, as well as its rescaling, $Z_j = \frac{\tilde{Z_j}}{c}$, is performed outside of the memristors crossbars. The non-linear activation function $f$ is applied: $f(Z_j)$. Finally, an average pooling is applied as $A_{i,j} = \operatorname{Avg}(f(Z))_{i,j} =  \frac{1}{s^2}\sum_{k=i}^{i+s}\sum_{l=j}^{j+s} f(Z_{k,l})$ where s is the kernel size.

%Given weight matrices $[w^{(1)},w^{(2)},\dots,w^{(P)}]$, input $x$,  and conductance range $[0,\Gmax]$:

%For each linear layer first, the conductance values are computed as $g = cw$ and then uniformly quantized in the conductance range $[0,\Gmax]$.

%Then the memristors products $\tilde{Z_j}^+ = \sum_{i=1}^L r G_{i,j}^+ x_i$ and $\tilde{Z_j}^- = \sum_{i=1}^L r G_{i,j}^- x_i$ are computed.

%Then, outside of the memristors crossbars, the difference are computed as
%$\tilde{Z_j} = \tilde{Z_j}^+ - \tilde{Z_j}^-$ and rescaled as $Z_j = \frac{\tilde{Z_j}}{c}$ 

%The non linear activation  function is applied: $X^{k+1}_j =  f(Z_j)$.

%The mean square error at the output of each layer is computed as 
%\begin{equation}
%    \operatorname{MSE}(f(z^p)) = \E[(f(z^p)-f(Z^p))^2] 
%\end{equation}

\section{Theoretical Analysis}
\label{sec:analytical}

\subsection{MSE prediction}

We now derive a theoretical analysis of the performance of a memristor-based implementation of neural network inference. As a proxy of performance, our goal is to predict the MSE between the noisy neural network outputs (computed using memristors) and the full precision (noiseless) outputs.
%\subsection{Matrix-vector multiplication}
We use the following notations throughout our analysis:
$\Var[G_{i,j}] = \sigma^2$,
$\E[X_{i}] = x_i$,
$\Var[X_{i}] = \gamma_i^2$,
$\Cov[X_{i},X_{j}] = \gamma_{i,j}$.

%First of all, it can be noted that since $G=G^+-G^-$,
The computation at the linear layer followed by the rescaling can be written as%, rewriting \eqref{eq:mvm_pos_neg} 
\begin{equation}
        Z_j = \frac{r}{c} \sum_{i=1}^L G_{i,j} X_i \,,
\end{equation}
and we can formulate the first and second moments of $Z_j$ as
\begin{equation}
\label{eq:mu_lin}
    \mu_j =  \E[{Z_j}]= r \sum_{i=1}^L ( w_{i,j}+\delta^q_{i,j})x_i \,,
\end{equation}
\begin{equation}
\label{eq:gamma_lin}
    \begin{split}
     \rho_j^2 = \Var[Z_j] =  r^2\Bigg(\sum_{i=1}^L \frac{\sigma^2 x_i^2}{c^2} +\gamma_i^2 (w_{i,j}+\delta^q_{i,j})^2+ \frac{\gamma_i^2 \sigma^2}{c^2} \\
            + \sum_{i=1}^L \sum_{i'=1,i'\neq i}^L ( w_{i,j}+\delta^q_{i,j})( w_{i'j}+\delta^q_{i',j}) \gamma_{i,i'}\Bigg) \,,
                \end{split}
\end{equation}
\begin{equation}
    \rho_{j,j'} =  \Cov[{Z_j},{Z_{j'}}] = r^2  \sum_{i=1}^L \sum_{i'=1}^L (w_{i,j}+\delta^q_{i,j})(w_{i'j'}+\delta^q_{i',j'}) \gamma_{i,i'} \,.
\end{equation}

%\subsection{Rescaling}

% Since the rescaling of the data is performed by dividing $\tilde{Z}_j$ by $c$, the moments may be recomputed as
% \begin{align}
%     \mu_j &=  \E[{Z_j}] = \frac{E[{\tilde{Z_j}}]}{c} \,, \label{eq:mu_lin}\\
%  \rho_j^2 &= \Var[Z_j] =  \frac{\Var[\tilde{{Z_j}}]}{c^2} \,, \label{eq:gamma_lin}\\
%   \rho_{j,j'} &=  \Cov[{Z_j},{Z_{j'}}] = \frac{\Cov[\tilde{{Z_j}},\tilde{{Z_{j'}}}] }{c^2} \,.
% \end{align}

%\subsection{Non-linear function}
%
Then, an approximation of the moments after the non-linear activation function $f$ is possible via Taylor expansions~\cite{dupraz_power-efficient_2021}:
\begin{align}
&\E[{f(Z_j)}] \approx f\left(\mu_{j}\right)+\frac{1}{2} f^{\prime \prime}\left(\mu_{j}\right) \rho_j^2  \, , \label{eq:mu_act}\\
&\Var[f(Z_j)] \approx \frac{1}{2} g^{\prime \prime}\left(\mu_{j}\right) \rho_{j}^{2}-f\left(\mu_{j}\right) f^{\prime \prime}\left(\mu_{j}\right) \rho_{j}^{2} \, ,\label{eq:gamma_act}\\
&\Cov[f(Z_j),f(Z_{j'})] \approx f^{\prime}\left(\mu_{j}\right) f^{\prime}\left(\mu_{j'}\right) \rho_{j, j'} \, ,
\end{align}
where $g=f^2$.
From these moments, the MSE of $f(Z_j)$ is
 \begin{equation}
     \operatorname{MSE}[f(Z_j)] = \Var[f(Z_j)]+(\E[{f(Z_j)}]-f(z_j))^2 \, .
 \end{equation}
The MSE can also be expressed as a function of $c$ as
\begin{equation}
      \operatorname{MSE}[f(Z_j)] = \frac{F_{1,j}}{c^4} + \frac{F_{2,j}}{c^2} +F_{3,j}
\end{equation}
The expressions of $F_{1,j}$, $F_{2,j}$, and $F_{3,j}$ can be computed by substituting~\eqref{eq:mu_lin} and~\eqref{eq:gamma_lin} in \eqref{eq:mu_act} and~\eqref{eq:gamma_act}.
Note that if $c \to \infty$, then $ \operatorname{MSE}[f(Z_j)] \to F_{3,j}$. Hence, $F_{3,j}$ gives us a lower bound on the MSE. Since this bound does not depend on $\sigma$, it is possible to find values of $c$ for any $\sigma$ that minimize the MSE to $F_{3,j}$.

%\subsection{Average pooling}

For average pooling (Section~\ref{sec:comp_model}), the moments are
\begin{align}
&\E[A_{i,j}] = \frac{1}{s^2}\sum_{k=i}^{i+s}\sum_{l=j}^{j+s} \mu_{k,l} \, ,\\
&\Var[A_{i,j}] = \frac{1}{s^4}\sum_{k=i}^{i+s}\sum_{l=j}^{j+s} \sum_{m=i}^{i+s}\sum_{n=j}^{j+s} \gamma_{k,l,m,n} \, ,\\
&\Cov[A_{i,j},A_{i',j'}] = \frac{1}{s^4}\sum_{k=i}^{i+s}\sum_{l=j}^{j+s} \sum_{m=i'}^{i'+s}\sum_{n=j'}^{j'+s} \gamma_{k,l,m,n}  \, .
\end{align}

\subsection{Power consumption}
\label{sec:power}

We now derive an estimation of the power consumption of the memristor computations. The power consumption of each memristor can be written as ${P^{\text{(mem)}}_{i,j}} = \left | G_{i,j} \right | X_i^2 \, ,$
%For each memristor ${P^{\text{(mem)}}_{i,j}}^+ = G_{i,j}^+ X_i^2$ and ${P^{\text{(mem)}}_{i,j}}^- = G_{i,j}^- X_i^2$ therefore we can write:
% \begin{equation}
%     {P^{\text{(mem)}}_{i,j}} = \left | G_{i,j} \right | X_i^2 \, ,
% \end{equation}
%
with
\begin{equation}
    \E[P^{\text{(mem)}}_{i,j}] = \E[G_{i,j} X_i^2] = c  \left |w_{i,j}+\delta^q_{i,j} \right |( \gamma_i^2+x_i^2) \, .
\end{equation}
Moreover, the power consumption of each transimpedance amplifier (TIA) is
\begin{equation}
     {P^{\text{(TIA)}}_{j}}^+ = \frac{(\sum_{i=1}^L  G_{i,j}^+ X_i)^2}{r} = \frac{\tilde{Z}_j^{+^2}}{r^2}
\end{equation}
and
\begin{equation}
     {P^{\text{(TIA)}}_{j}}^- = \frac{(\sum_{i=1}^L  G_{i,j}^- X_i)^2}{r} = \frac{\tilde{Z}_j^{-^2}}{r^2} \, , 
\end{equation}
with
\begin{equation}
     \E[{P^{\text{(TIA)}}_{j}}^+] = c^2 \frac{\rho_i^{+^2}+\mu_i^{+^2}}{r^2} \,, \quad
     \E[{P^{\text{(TIA)}}_{j}}^-] = c^2 \frac{\rho_i^{-^2}+\mu_i^{-^2}}{r^2} \, .
\end{equation}
Hence, the power consumption of each layer is
%\begin{equation}
%    \E[P_\text{tot}] = \sum_{j=1}^L(\sum_{i=1}^L \E[P^{\text{(mem)}}_{i,j}] +\E[{P^{\text{(TIA)}}_{j}}^+]+\E[{P^{\text{(TIA)}}_{j}}^-])
%\end{equation}
%\begin{equation}
%    \begin{split}
%\E[P_\text{tot}]        =\sum_{j=1}^L(\sum_{i=1}^L (c  \left |w_{i,j}+\delta^q_{i,j} \right |( \gamma_i^2+x_i^2)) \\+\frac{c^2}{r^2}(\rho_i^{+^2}+\mu_i^{+^2}+ \rho_i^{-^2}+\mu_i^{-^2}))    
%    \end{split}
%\end{equation}
\begin{equation}
\label{eq:E_Ptot}
    \E[P_\text{tot}] = \sum_{j=1}^L\Bigg(\sum_{i=1}^L \E[P^{\text{(mem)}}_{i,j}] +\E[{P^{\text{(TIA)}}_{j}}^+]+\E[{P^{\text{(TIA)}}_{j}}^-]\Bigg) \, .
\end{equation}

As a function of $c$, the power of each layer's column is
\begin{equation}
     \E[{P_\text{tot}}_j] =   c^2 H_{1,j}+ c H_{2,j}+H_{3,j} \,,
\end{equation}
where the expressions of $H_{1,j}$, $H_{2,j}$, and $H_{3,j}$ can be computed by developing the terms of equation~\eqref{eq:E_Ptot} from their definitions and equations~\eqref{eq:mu_lin} and~\eqref{eq:gamma_lin}.
\section{Optimization}
The parameter $\Gmax$ may be chosen with different granularity to balance design complexity and energy efficiency. For instance, one may apply the same $\Gmax$ to all memristor crossbars, associate a specific $\Gmax$ to each layer of the neural network, or even use a different $\Gmax$ per crossbar column.
We denote $\bm{\Gmax}$ as the set of $\Gmax$ variables that can be modified to optimize our computations. Depending if we have only one $\Gmax$ for the whole network or one $\Gmax$ for each layer, the size of $\bm{\Gmax}$ is $1$ or $P$, respectively.

To minimize the MSE for a specific power constraint, the global optimization problem can be formulated as 
%
%\begin{equation}
%     \label{eq:prob-1}
% \min_{\bm{\Gmax}}  \max{\operatorname{MSE}[f({Z^P})]}  \quad \text{subject to}  \quad \E[P_\text{tot}] \leq \mathcal{V} ,  \quad \Gmax^{(p)}>0 
%\end{equation}
\begin{equation}
     \label{eq:prob-1}
 \min_{\bm{\Gmax}}  \max{\operatorname{MSE}[f({Z^P})]} \, ,
\end{equation}
subject to $\E[P_\text{tot}] \leq \mathcal{V}$ and $\Gmax^{(p)}>0$. 
This corresponds to finding the best set of scaling constants $c$ that minimizes MSE for a desired total power usage.
The problem may be solved approximately using a heuristic optimizing search.

\section{Simulations}
\label{sec:results}
%goncalo: I am commenting the references but feel free to include them if we have space
We trained two convolutional neural networks on CIFAR-10 composed of five pairs of convolutional and average pooling layers and a final linear layer. Each subsequent convolutional layer in the smaller model has 2, 4, 8, 16, and 16 filters, as opposed to the 16, 32, 64, 128, and 128 filters of the larger model. We used a kernel size of 3 and a unit stride for all convolutional layers. 
%\jonathan{The last linear layer has 16 and 128 input features for the smaller and larger model, respectively.}
%The last linear layer of the smaller model has 16 neurons, whereas the larger model uses 128 neurons. 
%
Both models used the Softplus %~\cite{softplus} 
activation function and were trained for 164 epochs using stochastic gradient descent (SGD) with momentum, weight decay, and an initial learning rate of $0.1$ (decayed by $10$ at epochs 81 and 122). 
%As proposed by Hu \textit{et al.}~\cite{hu_dot-product_2016}, the input values were rescaled to between $0$ and $0.3V$.
The number of quantized values $N$ is set to $128$ and $r$ is set to $1$.

\subsection{Accuracy of the theoretical analysis}
\begin{figure}[t]
\centerline{\includegraphics[width=0.5\textwidth]{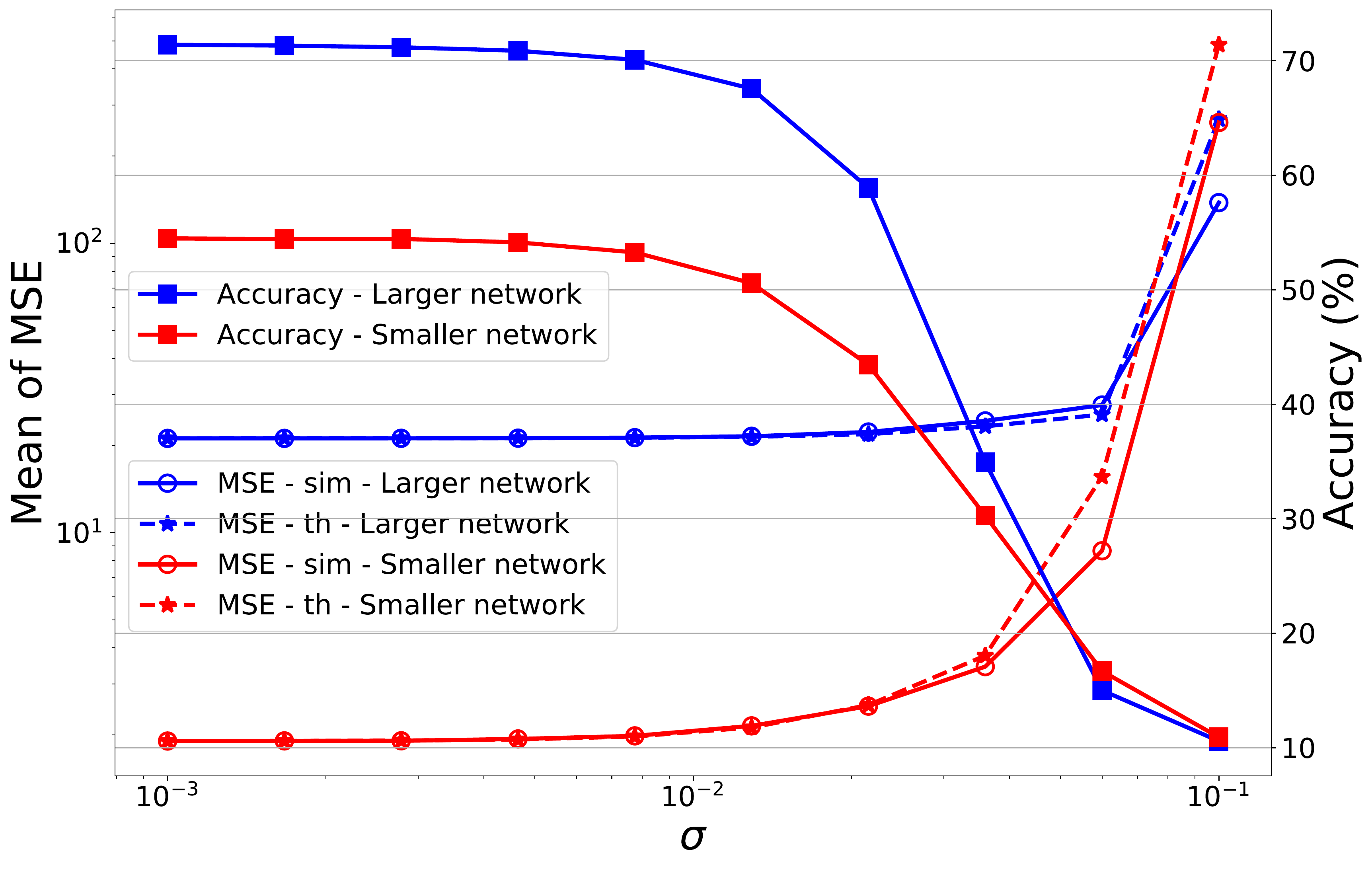}}
\caption{MSE at the output of the final layers of the smaller and larger network averaged over input examples, in terms of the standard deviation $\sigma$ of the conductance values.}
\label{fig:mse_for_sigma}
\end{figure}

Figure~\ref{fig:mse_for_sigma} shows the mean of the MSE on the final layer outputs of the smaller and larger models computed over 100 different inputs. These values are plotted both based on simulations and on the analytical formula presented in Section~\ref{sec:analytical}. We observe a close match between the theoretical and simulated MSE, especially in the high accuracy regime. Moreover, we see an inverse correlation between MSE and accuracy, which confirms predicting MSE to be a good proxy for estimating performance degradation. Moreover, as $\sigma$ decreases, the MSE converges to a value dependent on the quantization error.

With a Tesla P100 GPU, the mean runtime for the MSE computation of the small network on a batch of 64 inputs with $\sigma=0.01$ using our method is 27 ms. Under the same conditions, using 200 Monte-Carlo trials takes on average 2.3 seconds to reach a MSE within 2\% of the true MSE 98\% of the time. This 85$\times$ speedup showcases the usefulness of our method in practice. 
%It should be noted that computing the theoretical MSE took around a tenth of time required for computing the simulated MSE. 
%This shows the usefulness of the theoretical analysis for predicting  the performance of the network given a set of parameters.

\begin{figure}[t]
\centerline{\includegraphics[width=0.5\textwidth]{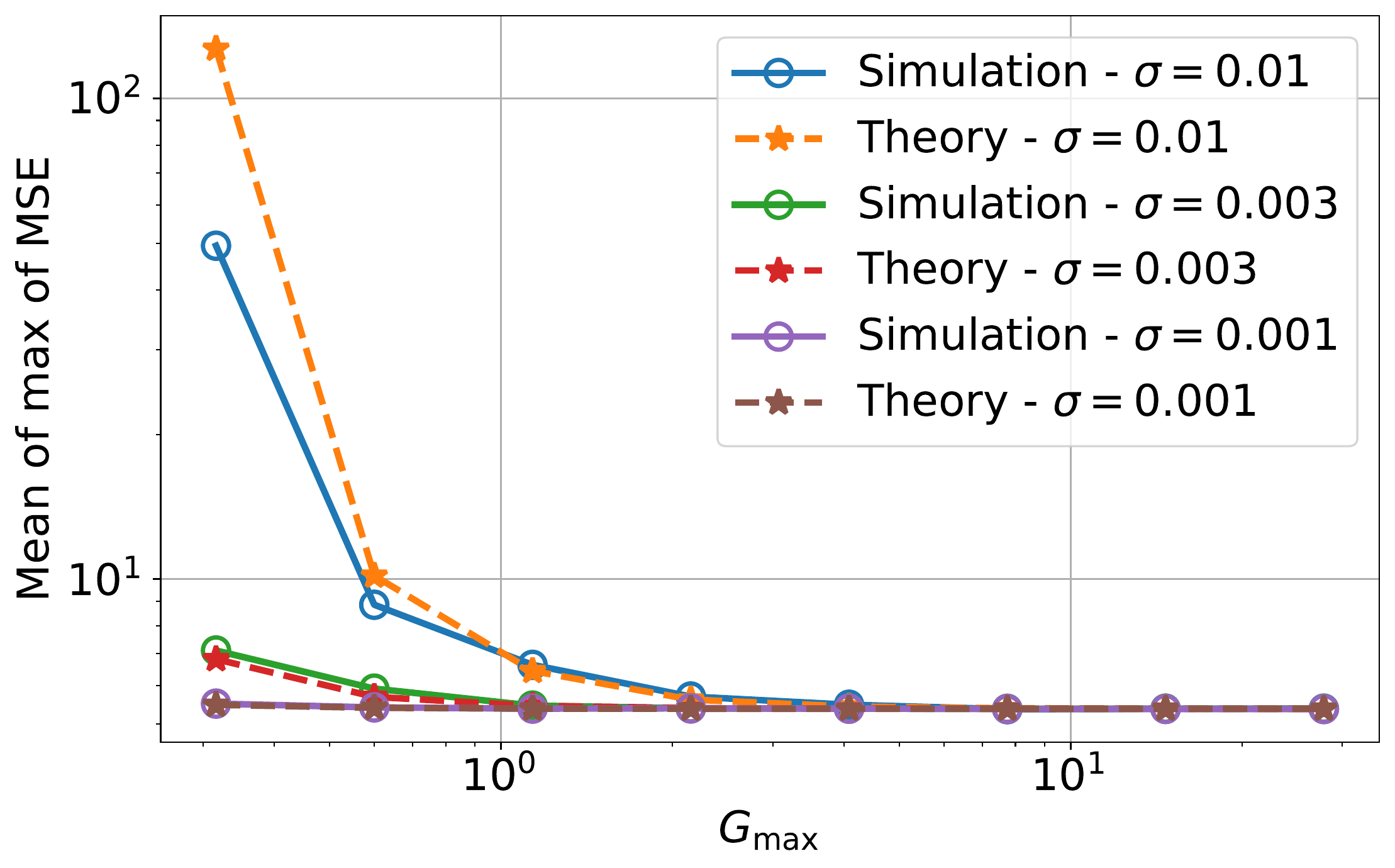}}
\caption{Mean of the maximum of the MSE of the output of the smaller network's final layer depending on $\Gmax$ and $\sigma$ values.}
\label{fig:mse_for_gmax}
\end{figure}

Figure~\ref{fig:mse_for_gmax} shows the mean of the maximum MSE on the output of the smaller model depending on the value of $\Gmax$, for different values of $\sigma$. Once again, we observe that the theoretical computations accurately predict the simulation results. Moreover, we see the predicted convergence to the theoretical bound. 
Such bound is reached faster as $\sigma$ decreases. 
In Figures~\ref{fig:mse_for_sigma} and~\ref{fig:mse_for_gmax}, we notice that for a high ratio of noise to $\Gmax$ there is a gap between theoretical and simulation results. This is likely due to the Taylor expansions used for approximating the moments after the activation function.

\subsection{Numerical optimization}

\begin{figure}[t]
    \centerline{\includegraphics[width=0.5\textwidth]{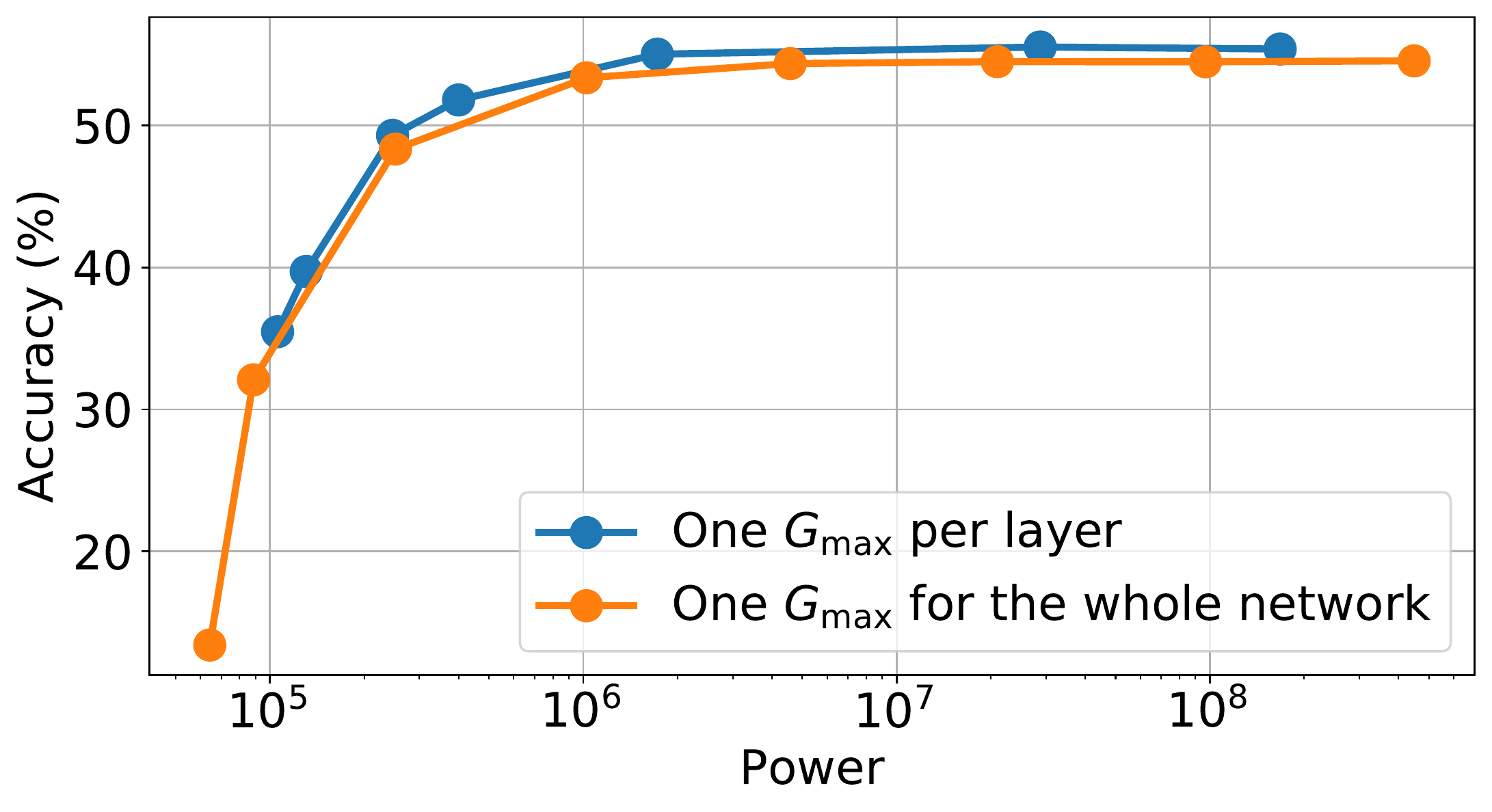}}
    \caption{Maximization of the smaller network accuracy using a genetic algorithm minimizing the maximum of the MSE for different power constraints with $\sigma=0.01$}
    \label{fig:opti_power_for_mse}
\end{figure}

Figure~\ref{fig:opti_power_for_mse} shows the results of optimizing $\bm{\Gmax}$ following \eqref{eq:prob-1} for the smaller network. For each power constraint, a genetic optimizer was run for 100 generations with a population size of 50 and a sample of 100 inputs for computing the mean of the theoretical MSE and power consumption of the network. The proposed approach allows to efficiently find the value(s) of $\Gmax$ that minimize MSE (maximize accuracy) for a given power constraint. As expected, adding degrees of freedom by allowing a different $\Gmax$ for each layer leads to improved performance, although the benefit is marginal in this case.

%Finally, we use global optimization for solving problem~\eqref{eq:prob-2}. For this problem we want to minimize the MSE of all outputs of the final layer given a maximum power budget. We are therefore in the case of multi-objective optimization we require the use of special optimization algorithms such as NSGA-II~\cite{deb_fast_2002}. In this case, the algorithm will return an array of possible solutions and we can choose from this array the solution that best corresponds to the application's criteria. Here, we choose the solution with the lowest maximum of the output MSE vector. 
%The results can be seen in Figure~\ref{fig:opti_mse_for_power}.

% \begin{figure}[t]
%     \centerline{\includegraphics[width=0.5  \textwidth]{pymoo_NSGAII_sigma001_comp_minimize_mse_multi_objectives_minmax.pdf}}
%     \caption{Multi objective minimization of the mean MSE using NSGA-II for a given power budget with $\sigma=0.01$}
%     \label{fig:opti_mse_for_power}
% \end{figure}

\section{Conclusion}

In this work, we studied the implementation of DNN models using memristors crossbars. Using second-degree Taylor expansions, we proposed approximate theoretical formulas of the MSE at the output of the network, as well as theoretical computations of the power usage of the memristors. 
We then considered
%Using these analyses, it is possible to express 
an optimization problem for maximizing task performance under a power usage constraint.
%for minimizing the power usage for a desired performance of a neural network. 
The theoretical analysis makes it feasible to solve this optimization problem numerically since its computing time is faster than using simulations by almost two orders of magnitude.

\bibliographystyle{ieeetr}
%\bibliography{test}
\bibliography{refs}

\end{document}